\documentclass[10pt,runningheads]{llncs}
\usepackage[margin=1in]{geometry}
\usepackage{tabularx,booktabs}
\usepackage{url}
\usepackage{times}
\usepackage{graphicx}
\usepackage{lipsum}
\newcommand{\junk}[1]{}
\usepackage{ulem}
\usepackage{multirow}
\usepackage{booktabs}
\usepackage{color}
\usepackage{subcaption}
\usepackage{amsfonts}
\usepackage{amsmath}
\usepackage[dvipsnames]{xcolor}
\usepackage{tikz}
\usepackage{threeparttable}
\usepackage{caption}
\usepackage{lineno}
\usepackage{float} 
\linenumbers
\usepackage{cite}
\usepackage{authblk}
\usepackage[numbers,sort&compress]{natbib}
\makeatletter
\def\blfootnote{\xdef\@thefnmark{}\@footnotetext}
\usepackage{hyperref}

\usepackage{xr}
\makeatother

\begin{document}
\nolinenumbers

\title{Drug Synergistic Combinations Predictions via Large-Scale Pre-Training and Graph Structure Learning}
\renewcommand{\thefootnote}{\fnsymbol{footnote}}

\author{Zhihang Hu\footnote[0$^\dagger$]{$^\dagger$ Equal first authorship.}\inst{1} \and Qinze Yu$^\dagger$\inst{1} \and Yucheng Guo\inst{2} \and Taifeng Wang\inst{2} \and Irwin King\inst{1} \and Xin Gao$^\ast$\inst{3} \and Le Song\footnote[0$^\ast$]{$^\ast$ Corresponding Author. 
}\inst{2} \and Yu Li$^\ast$\inst{1,4}}
\authorrunning{Z. Hu et al.}

\institute{ Department of Computer Science and Engineering, The Chinese University of Hong Kong, Hong Kong SAR, China \and
BioMap (Beijing) Intelligence Technology Limited, Beijing, China\and Computational Bioscience Reseach Center, King Abdullah University of Science and Technology, Thuwal, Saudi Arabia\and The CUHK Shenzhen Research Institute, Hi-Tech Park, Nanshan, Shenzhen, China}
\maketitle 
\begin{abstract}
Drug combination therapy is a well-established strategy for disease treatment with better effectiveness and less safety degradation. However, identifying novel drug combinations through wet-lab experiments is resource intensive due to the vast combinatorial search space. Recently, computational approaches, specifically deep learning models have emerged as an efficient way to discover synergistic combinations. While previous methods reported fair performance, their models usually do not take advantage of multi-modal data and they are unable to handle new drugs or cell lines. In this study, we collected data from various datasets covering various drug related aspects. Then, we take advantage of large-scale pre-training models to generate informative representations and features for drugs, proteins, and diseases. Based on that, a message-passing graph is built on top to propagate information together with graph structure learning flexibility. This is first introduced in the biological networks and enables us to generate pseudo-relations in the graph. Our framework achieves the state-of-the-art results in comparison with other deep learning-based methods on synergistic prediction benchmark datasets. We are also capable of inferencing new drug combination data in a test on an independent set released by AstraZeneca, where 10\% of improvement over previous methods is observed. In addition, we're robust against unseen drugs and surpass almost 15\% AU ROC compared to the second best model. We believe our framework contributes to both the future wet-lab discovery of novel drugs and the building of promising guidance for precise combination medicine.
\end{abstract}
\newpage
\section{Introduction}
Drug combination therapy has been widely applied in both traditional and modern medicine due to its diverse merits. Compared with monotherapy, administering drug combinations leads to improvement of efficacy \cite{csermely2013structure}, and reduction of side effects \cite{zhao2013systems} and host toxicity \cite{o2016unbiased}, further, it even overcomes drug resistance \cite{hill2013genetic}. Considering the fact that a single drug usually cannot be effective, drug combinations are increasingly used to treat a variety of complex diseases, such as human immunodeficiency virus (HIV) \cite{clercq2007design}, virus infections \cite{zheng2018drug}, and cancer \cite{kim2021anticancer, al2012combinatorial}. For instance, the combination of two clinically used drugs, colloidal bismuth subcitrate (CBS) and N-acetyl cysteine (NAC), suppresses the replication cycle of SARS-CoV-2 virus and reduces viral loads in the lung \cite{wang2022orally}. The combination provides a potential treatment for combating SARS-CoV-2, which can hardly be treated by any single drug. However, drug combinations can also be harmful without precise medicine \cite{hecht2009randomized, azam2021trends}. Therefore, it is pretty important to accurately find synergistic drug pairs for a cell type in case we want to take advantage of drug combination therapy.


Traditional methods for drug combination discovery are mainly based on clinical trial and error, which is time- and cost-consuming and can result in harm to patients \cite{day2016approaches, pang2014combinatorial}. Besides, the limited resources only satisfy web-lab tests on a few drug combinations \cite{li2015large}. 
With the development of experimental technology, researchers are able to carry out high-throughput drug screening (HTS) \cite{macarron2011impact, torres2013high, he2018methods}, which is a kind of sensitive and fast synchronous experiment and makes the exploration of large drug combination space become a reality. 
Due to HTS, the drug combination synergy data had increased tremendously. Some public databases make contributions to drug research for specific tissues, like ASDCD \cite{chen2014asdcd} provides antifungal drug combinations data, and a large HTS synergy study \cite{o2016unbiased} performed more than 20000 drug synergy measurements, which covers 38 drugs and 39 cancer cell lines. Part of these databases offer high quality training data for the development of computational methods, and also help the evaluation of computational methods for predicting novel drug combinations.
However, the in vivo and in vitro experiments cannot be exactly consistent. Although the original tumor and the derived cancer cell line share a high degree of genomic correlation, in vitro experiments are not able to restore the mode of drug action in vivo \cite{ferreira2013importance}, which means there still exist obstacles impede the effectiveness of HTS.

In recent years, with the advance of computational technology, some machine learning models and neural networks are effective and promising in finding novel drug combination candidates in large synergistic space. For example, DeepSynergy \cite{preuer2018deepsynergy} combined the three different types of chemical features of drugs and genomic information of cancer cells to predict drug pairs with synergistic effects. TranSynergy \cite{liu2021transynergy} is a transformer-based method that integrates information from gene-gene interaction networks, gene dependencies, and drug-target associations to predict synergistic drug combinations and deconvolute the cellular mechanisms. DeepDDS \cite{wang2022deepdds} converts molecular drugs into graphs and proposes a graph neural network with an attention mechanism to identify the synergistic drug combinations. MR-GNN \cite{xu2019mr} extracts features from different neighborhoods of each node in the drug molecular graph and uses a dual graph-state LSTM-based network to process features.
Some other methods like DeepDDI \cite{ryu2018deep}, DeepDrug \cite{yin2022deepdrug}, and GCN-BMP \cite{chen2020gcn}, focus on resolving drug-related tasks like drug-drug interaction, drug-food interaction, and drug relations, which provides useful information for synergistic drug combination prediction task.
Nevertheless, their prediction target usually remains in a specific pathway, cell line, or tissue because of the limitation of their used dataset. These studies are usually based on single databases, and the prediction was also made within the database. If we want to develop an unbiased and generalizable drug synergy prediction model, one of the key challenges is the problem of domain-shift data: the invitro drug responses of different tissues can be various. Previous methods focus on studying common tissues, like breast, skin, and lungs \cite{tang2019network, chen2018predict, zhao2011prediction, huang2014drugcomboranker}, these methods use drug combinations data of certain cell lines for training and try to discover novel drug combinations from other cell lines within the same tissue. Therefore, some tissues remain understudied due to some difficulties in data or bio-experiments. For instance, bone cancer is hard to deal with because of the technical limitations in culturing bone tissue as cell lines. The lack of cell line models leads to the obstacle in high-throughput screening, which in turn makes these tissues more difficult to study. As a consequence, finding out a way to develop a generalized drug combination effect prediction model is essential for resolving understudied tissues problem. Fortunately, with the help of the databases that systematically integrate multiple drug synergy datasets \cite{zagidullin2019drugcomb, liu2020drugcombdb}, the roadway to developing unbiased drug synergy prediction models was not that elusive.


In this study
, we address the problems mentioned above by proposing an end-to-end deep learning framework that accurately predicts synergistic effects. Our method takes advantage of multi-modal data, graph neural networks, and large-scale unsupervised training to integrate and learn useful information for synergistic prediction. Specifically, our model takes chemical structure graphs of drugs and the protein expression of cell lines as input and applies a pre-trained molecular graph transformer \cite{li2022kpgt} to convert drug graphs into embeddings. Meanwhile, the model generates embeddings for every protein in the expression by utilizing a protein language model \cite{rives2021biological}. To enrich more features, we also include disease information, particularly, we apply RotatE \cite{sun2019rotate} to get the embedding of disease from PrimeKG \cite{chandak2022building}. Next, we utilize graph neural networks and take our generated embeddings as node representations. In order to inference on unseen drugs, we include drug-drug similarity edge and drug-target module/ drug-drug interaction module to generate pseudo edges and formed a refined graph with richer information. Finally, a synergistic prediction head is built on top of our graph and  acts as a Perceptron (MLP) to predict the synergistic effect. We also include a self-training strategy to make use of the large amount of data in combination space. 

We compare our model with five task-related deep learning models and two traditional machine learning models on the benchmark datasets, DrugComb and AstraZeneca. Our framework outperforms all the previous methods on various evaluation criteria.  Moreover, we conduct experiments on some unseen drugs and cell lines in the training set to justify our robustness. We believe that our method is an effective tool for discovering novel synergistic drug combinations for further wet-lab experiment validation.

\section{Methodology}
In this section, we will dive into details of the dataset construction and how we establish our training pipeline. We divided this part into several subsections: Preprocessing section describes datasets manipulation and feature pre-training; Heterogeneous graph section delivers  graph construction, graph neural network, and synergistic prediction head information; Graph structure learning section introduces our Drug-Target predictive module, Drug-Drug interaction module, and graph structure learning details; Self-training and inference section summarizes our self-training strategy and the way to perform inference.

\begin{figure}[!t]
\centering
\includegraphics[width=1\textwidth]{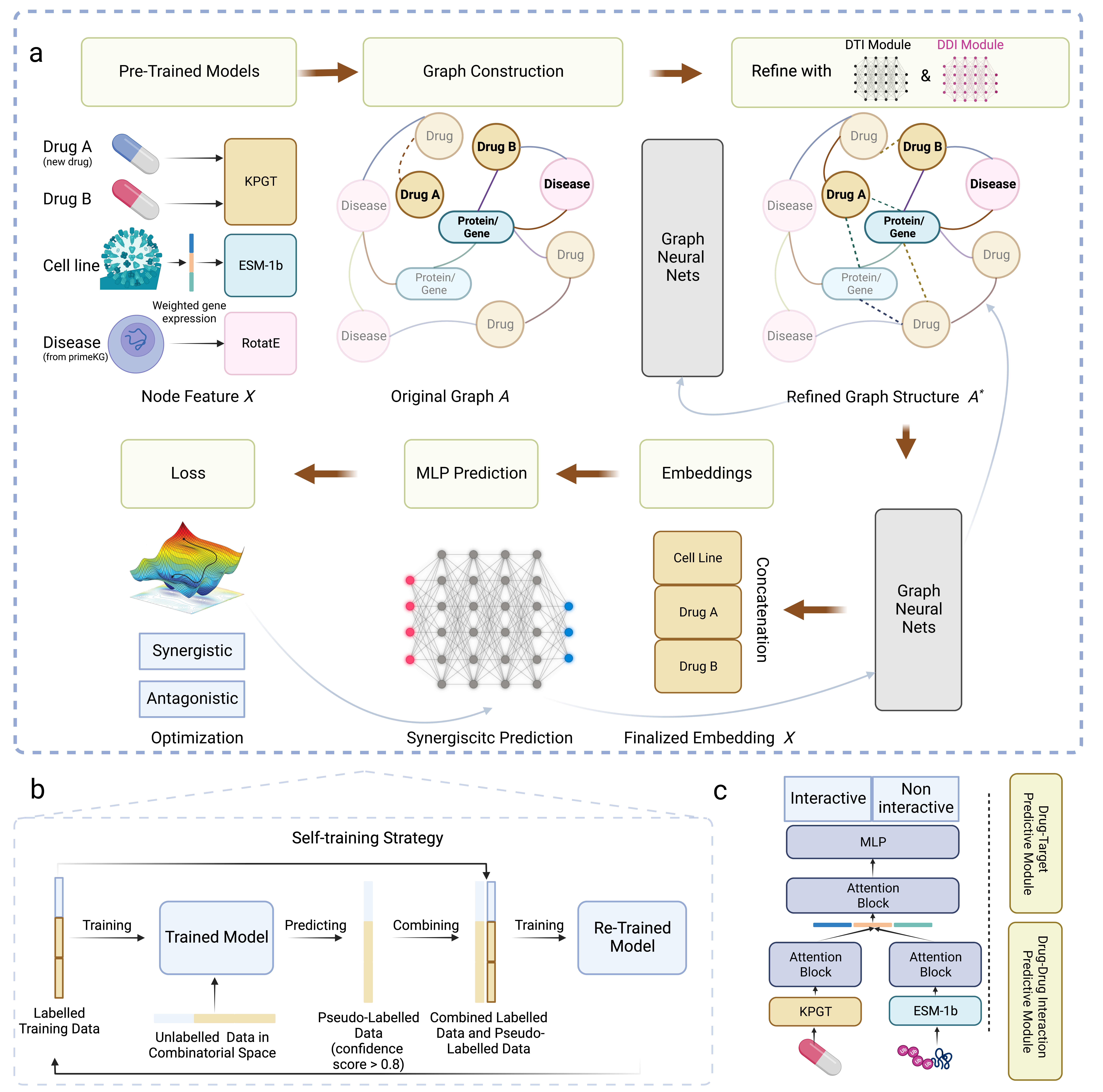} 
\caption{
\textbf{Overview of our drug synergistic combinations prediction framework.} \textbf{a.} We obtain features of drug, protein, and disease from three pre-trained models respectively, and build a heterogeneous graph upon these node features. Next, we perform drug-drug and drug-target inference to propagate information and refine the graph. The finalized embeddings of drugs and the cell line are obtained through a graph neural network based on the refined graph. The embedding vectors are subsequently concatenated to feed into a Multilayer Perception (MLP) to predict the drug combination synergy effect. \textbf{b.}Details of our self-training strategy to make use of large unlabelled synergistic data. \textbf{c.} The architecture of our Drug-Target predictive module,  and our Drug-Drug interaction predictive module is built in the same way.}
\label{Fig.overview}
\end{figure}

\subsection{Preprocessing}
We take advantage of various types of datasets to meet our requirements for diverse information, in this section, we introduce each of the used datasets in detail by the order of our framework pipeline. 

\paragraph{Datasets}
The datasets we used include originally published datasets as well as some following databases which integrated them together. They covered multiple aspects and biological relations, here we listed them below,\begin{itemize}
    \item \textbf{\textit{Zinc250k\cite{irwin2005zinc}/ChemBL}}: Large scale drug SMILES dataset used for unsupervised training. 
    \item \textbf{\textit{PrimeKG\cite{chandak2022building}}}: A knowledge graph dataset integrating 20 high-quality datasets, biorepositories, and ontologies.
    \item \textbf{\textit{Therapeutics Data Commons\cite{huang2021therapeutics}}}: TDC is a resource platform to access curated AI-ready datasets, machine learning tasks, and benchmark datasets.
    \item \textbf{\textit{DrugComb\cite{zagidullin2019drugcomb}}}: DrugComb is an open-access, community-driven data portal where the results of drug combination screening studies for a large variety of cancer cell lines.
    \item \textbf{\textit{AstraZeneca}}: AstraZeneca is an independent dataset published by AstraZeneca company and involves several drug combination screening studies on multiple cancer cell lines.
    
\end{itemize}

After we obtained the above datasets, we need to combine equivalent entries and delete duplicates. \begin{itemize}
    \item Drug data are unified by their DrugBankID or ChemBLID across all our datasets and their corresponding SMILES representations are unified to their canonical form.
    \item Protein data in PrimeKG and TDC are aligned via their Gene ID, Uniprot ID, and Protein ID.
    \item Our cell line data are originally expressed on 13418 proteins obtained via CCLE dataset. After examining existing protein data in PrimeKG and TDC, we delete 10 protein columns in our cell expression.
\end{itemize}

Finally, after pre-processing, we can obtain around 22,000 numbers of protein/genes, 2000 Drugs, and 15,000 Diseases. They can be treated as nodes in our graph neural network but we still lack an important ingredient - their representations. We gained their representations through large-scale pre-training methods explained below.

\paragraph{Feature Representation}

Unsupervised learning shows brilliant results in the field of Natural Language Processing and bioinformatics, many sequence-based biological tasks benefit from it. Inspired by the progress of unsupervised frameworks, we take advantage of three models with large-scale information: ESM-1b \cite{rives2021biological}, KPGT \cite{li2022kpgt}, and PrimeKG \cite{chandak2022building} to obtain the initial embedding of our protein, drug, and disease respectively. 

\textbf{\textit{Protein: ESM-1b.}}
ESM-1b is trained on Uniprot \cite{uniprot2015uniprot} database, utilizing 250 million protein sequences, which is large enough to
support the high-capacity protein language model. And deep transformer is chosen as the architecture, for its
great performance in many tasks. During the training process, the amino acid sequences extracted from database are further divided into different fractions and given a special mask token as inputs of model, and the output of neural network is the missing token of corresponding sequences. With the pretrained ESM-1b, we input a protein sequence and get residue-level sequence embeddings, and then average across all residue positions of such embeddings, so that we finally get a $768$-dimension feature for the input sequence. 

\textbf{\textit{Drug: KPGT.}}
KPGT is trained on two million molecular SMILES from ChEMBL29 dataset \cite{gaulton2017chembl} for the representation learning of molecular graphs. With the Line Graph Transformer (LiGhT) structure, the model focuses on chemical bonds and can capture the structural information of molecular graphs. And the knowledge-guided pre-training strategy helps to exploit the additional message like abundant structural and semantic information, which are potentially more important to the downstream tasks, inside the molecular graphs. Given a SMILES string, the KPGT can convert it into a molecular line graph and input it to LiGhT. At last, we can get the $2304$-dimension embedding of our input SMILES string.

\textbf{\textit{Disease: RotatE.}}
Precision Medicine Knowledge Graph (PrimeKG) presents a holistic view of biological factors including diseases. To generate disease embeddings, we apply RotatE \cite{sun2019rotate} on top of PrimeKG to learn and gather the information. In our task, RotatE defines the disease relation as a rotation from the source node to the target node in the complex vector space, and the method is effective in modeling three relation patterns: symmetric/antisymmetric, inversion, and composition, therefore, it is able to resolve all relations in PrimeKG. Finally, we obtain a $512$-dimension embedding for each disease node.

\textbf{\textit{Cell line: Depmap CCLE}}
DepMap provides a “cancer dependency map” CCLE dataset by systematically identifying genetic dependencies. Our cell line data retrieved from CCLE are expressed on around 13400 genes/proteins. After preprocessing, a cell line therefore can be represented by a 13400-dimension vector, with each entry referring to the value of the corresponding gene/protein. We can then take advantage of our pre-trained protein embeddings and replace our original cell representation with the summation of weighted protein embeddings as shown below:
\begin{eqnarray}
    h_{j} = \sum_{i \in protein} CCLE_{j}[i] \cdot E_i, j\in Cell,
\end{eqnarray}

where $CCLE_{j}$ is the CCLE 13400 dimension gene/protein expression vector of a cell $j$, $E_{i}$ corresponds to the $768$ dimension ESM-1b embedding for protein $i$, and $h_j$ is our derived cell line embedding. In this way, we encode richer information to our new cell line representations considering both gene expression and gene structural/co-evolutionary information.

\subsection{Heterogeneous Graph}
As shown in Figure \ref{Fig.overview}, after we obtain the initial representation for each node, we then construct the heterogeneous graph based on relations extracted from PrimeKG and TDC. In total, there are nine edge types and three node types in our graph. We listed all edge types in Table \ref{graph} including their connecting nodes and descriptions. Note that all edge types except Drug-Drug Similarity can be inferred in our datasets directly, we include Drug-Drug Similarity to benefit information propagation in our graph, especially for unseen drugs. Drug-Drug Similarity here is computed by measuring the distance between our KPGT drug embeddings as well as fingerprints Tanimoto similarity, if the distance between two drugs is smaller than $90$ or their similarity greater than $0.62$, we connect these drugs with Drug-Drug Similarity edge. Finally, with these relation types and nodes, we can construct our heterogeneous graph $G = (A,X)$ where $A,X$ here stands for the initial adjacency matrix and node embeddings.

\textbf{\textit{Graph neural networks}} $\psi$ are built upon our constructed $G$ to conduct message passing. Here in our framework, a simple Multilayer perception module and a Graph attention architecture are applied.  We noticed that this simple architecture is adequate  to reach similar performance \cite{10.1145/3447548.3467350} compared to computation demanding heterogeneous GNN framework such as MAGNN, graph-transformer, etc. Each node embeddings $\{Drug: 2304,\: Protein: 768, \: Disease:512 \}$ propagate through different MLP modules to form unified vectors with length $512$. Take protein embeddings as an example, an MLP takes 768-dimension vectors as input and outputs 512-dimension vectors for network propagation. These updated vectors with the same length of each instance served as
their new embeddings, and then these embeddings proceed through a homogeneous graph attention network to propagate
information as shown in Figure~\ref{Fig.overview}. We inserted three GNN layers in our framework. One is directly after the initial graph construction and computes upon the adjacent matrix $A$. Another two are after our graph refinement and compute upon the refined adjacent matrix $A^{*}$.

\textbf{\textit{Synergistic Prediction Head}} $f_{syn}$ is located at the last part of our framework. With the final embedding vectors of our graph node $X^{*}$, we can extract our final drug representations as well as the cell line representations being the weighted sum of its expressed genes. Then two drug embedding vectors and one cell line vector are sent through MLP, they are concatenated as the input of multiple fully-connected layers. The output of this MLP module is two prediction scores for synergistic or antagonistic classes. The label of the synergistic effect was computed by the softmax function that follows the output of prediction scores.

\begin{table}[!t]
\centering
\caption{Nodes and edges in our heterogeneous graph}
\label{graph}

\scalebox{1}{
\begin{tabular}{cccccc}  
\toprule
Relation & Node & Node & Description \\
\midrule
 Drug-Drug Interaction P  & Drug   & Drug  &Summarized Drug-Drug effect obtained from DrugBank \\ 
 Drug-Drug Interaction N  & Drug   & Drug  &Summarized Drug-Drug effect obtained from DrugBank \\ 
 Drug-Drug Similarity  & Drug   & Drug  &Drug similarity based on fingerprints and KPGT.\\
  Drug-Target Interaction  & Drug   & Protein/gene  &Drug target interaction on specific proteins.\\
Drug-Disease Indication  & Drug   & Disease  &Drug Disease interaction based on effect. \\ 
 Drug-Disease Contraindication  & Drug &Disease  &Drug Disease interaction based on effect.\\
  Protein-Protein  & Protein/Gene  & Protein/Gene  &Protein protein interaction obtained from TDC and PrimeKG\\ 
  Protein-Disease  & Protein/Gene  & Disease  &Where disease is related with specific protein/gene.\\ 
  Disease-Disease  & Disease  & Disease  &Disease Disease interaction.\\ 

 \bottomrule
 \end{tabular}}
\end{table}

\subsection{Graph Structure Learning}
Graph structure learning(GSL) has emerged as a new technique for learning adaptive graphs when only insufficient data about the graph is available. Here, we introduce GSL into our framework in order to learn new pseudo edges to conduct inductive inference for  unseen drugs during testing. Moreover, experimental results indicate that even the prediction accuracy of current nodes will increase by adding pseudo edges since it benefits information propagation. While there are 9 types of edges in our graph, Drug-Target interaction and Drug-Drug interaction are the two most informative edge types for unseen drugs. The idea of our graph structure learning is thus built upon a predictive module which consists of a Drug-Target interaction module and a Drug-Drug interaction network to generate pseudo edges. We will illustrate how these two modules are constructed and trained below.

\textbf{\textit{Drug-Target Interaction Module.}}
The most informative relation in our graph lies in the Drug-Target edges since they update the drug and cell line embeddings directly. Our Drug-Target interaction (DTI) module is pre-trained separately and finetuned into our framework. The data we utilized for DTI training are collected via PrimeKG and TDC which originates from BindingDB \cite{liu2007bindingdb}. The data pairs are in $(Drug, Protein)$ and form existing interaction. Thus, we trained an MLP classifier that takes two embeddings as input and outputs the  prediction score. The DTI module architecture is presented in Figure \ref{Fig.overview}, it is a simple stack of several attention blocks and multiple perceptron layers. KPGT drug embeddings and ESM-1b protein representations go through separate computation branches and concatenate together through a predictive head. We train our DTI module with all positive existing data pairs and run negative sampling three times the number of positive pairs. The negative sampling factor can be modified, and we set it to be larger than one since we tend to control the numbers of predicted pseudo edges not to explode.

\textbf{\textit{Drug-Drug Interaction Module.}} Another informative relation in our graph is Drug-Drug interaction (DDI) edges since they also directly update the drug embeddings. Our Drug-Drug interaction module is also pre-trained separately and fine-tuned into our framework. The data we utilized for DDI training are collected via PrimeKG and TDC which originates from DrugBank and Twosides \cite{tatonetti2012data}. The data pairs are in $(Drug, Drug)$ form with $P$ or $N$ labels summarizing the interaction effect. The architecture is presented in Figure~\ref{Fig.overview}. Likewise, we trained an MLP classifier that takes two embeddings as input and outputs the predicted category $(P,N,$no edge). The DDI module architecture and training strategy are similar to DTI.

\textbf{\textit{Graph Structure Learning.}} After our DTI and DDI predictive module is pre-trained. We can fit them into our framework to generate pseudo edges and tune our framework end to end. Suppose our original heterogeneous graph $G$ can be presented by its adjacency matrix $A$, then the refined graph $A^{*}$ can be gained via our predictive module, $A^{*} = g(f_{DTI}, f_{DDI}, A)$, $g$ here act like a $sgn$ function,

\begin{eqnarray}
            A^{*}_{ij} = g(f_{DTI}, f_{DDI},A ) = \left\{
\begin{aligned}
f_{DTI}(X_{i},X_{j}) & , & e_{ij} = e_{DTI} \: and \: A_{ij}=0, \\
f_{DDI}(X_{i},X_{j}) & , & e_{ij} = e_{DDI} \: and \: A_{ij}=0, \\
A_{ij}& , & Otherwise
\end{aligned}
\right.
\end{eqnarray}
Finally, the refined graph passes through a graph neural network and enters our synergistic prediction head. We denote the initial embedding for each node as $X$, and the finalized embedding as $X^{*}$. Thus, $X^{*} = \psi(A^{*}, X)$. The whole framework is then tuned according to the loss below:
\begin{eqnarray}
            L &= &BCE(Y,f_{syn}(X^*))= BCE(Y,f_{syn}(\psi(A^{*}, X))\\\nonumber
            &= &BCE(Y,f_{syn}(\psi(g(f_{DTI}, f_{DDI}, A), X)) \\ \nonumber
            min\: L &=& \min_{f_{syn},\psi,f_{DTI}, f_{DDI}} BCE(Y,f_{syn}(\psi(g(f_{DTI}, f_{DDI}, A), X))
\end{eqnarray}

$Y$ here refers to the ground truth synergistic labels and $\psi$ stands for graph neural network which propagates messages on $A$ and $A^{*}$. $BCE$ stands for the binary cross entropy function to calculate our corresponding classifcation loss. Thus, the submodules that can be optimized in our framework include our synergistic prediction module $f_{syn}$, graph neural network $\psi$ and pre-trained predictive module $f_{DTI}, f_{DDI}$.

\subsection{Self-training and inference}
Self-training has shown positive effects in limited-data supervised learning tasks. Here in our case, although Drug-Comb provides over 300,000 entries, the combinatorial search space actually consists of over 0.7 billion possible cases. Labeled data doesn't even occupy $0.1\%$ of the whole space. Thus, we believe by expanding training data through self-training, our performance can be certainly boosted forward. Our main idea lies in using predicted confidence scores to filter out new training data. Figure \ref{Fig.overview} visualizes this procedure in our framework. First, we train our model on the original dataset $S$. Then, we run inference on the 0.7 billion combinatorial search space and obtained those entries $U$ whose confidence scores are greater than 0.8, we controlled the number of $U$ to be smaller than our original dataset $S$ and merged them together: $S^{'} = S\cup U$, to make a new training set. We then retrained our model on $S^{'}$. This process converges until our re-trained model almost cannot gain improvement.

After our framework is trained, for an incoming triple $(Drug_A,Drug_B,Cell\_line)$,  inference can be conducted easily on our method. \begin{itemize}
    \item First, generate both drug embeddings for $Drug_A,Drug_B$.
    \item  Inspect whether $Drug_A,Drug_B$ are in our graph. If yes, remained unchanged else let $Drug_A$ to be an unseen drug. Involve $Drug_A$ into our graph $G$ and generate Drug-Drug similarity edge for $Drug_A$ and obtain graph $G'$.
    \item  Run GNN on $G$ or $G'$, conduct Drug-Target, Drug-Drug interaction inference and generate pseudo edges with refined graph $G^{*}$.
    \item  Run GNN on $G^{*}$, and gained the finalized embedding $X^{*}$
    \item  Conduct synergistic prediction based on $f_{syn}$ and $X^{*}$.
\end{itemize}

\section{Results}

\begin{table}[!t]
\centering

\caption{Performance comparison of 10-fold cross-validation on DrugComb dataset.}
\label{test_set}
\scalebox{1.1}{
\begin{threeparttable}

\begin{tabular}{p{2.6cm}|p{1.5cm}|p{1.5cm}|p{1.5cm}|p{1.5cm}|p{1.5cm}|p{1.5cm}}
\toprule
\textbf{Metric} & \textit{AU ROC} & \textit{AU PRC} & \textit{ACC} & \textit{BACC} &\textit{Precision}\tnote{a} &\textit{F1-Score} \\
\midrule
 
 \textbf{Ours}   & \textbf{0.961}  & \textbf{0.954} &0.878   & \textbf{0.862} & 0.883 & \textbf{0.972} \\
 DeepDDS\cite{wang2022deepdds}  & 0.942   & 0.934 &0.865 &0.855 &  0.862 &0.957         \\ 
 TranSynergy\cite{liu2021transynergy}  & 0.912  & 0.918 &0.892 &0.814 & 0.845&0.924        \\
 DeepSynergy\cite{preuer2018deepsynergy}  & 0.894  & 0.882 & \textbf{0.894} &0.862 & 0.843 &0.894 \\
 MR-GNN\cite{xu2019mr}  & 0.935  & 0.917 &0.885 &0.891 & \textbf{0.916} &0.904        \\
 MatchMaker\cite{kuru2021matchmaker}  & 0.927  & 0.914 &0.853 &0.876 & 0.786 &0.893        \\
 XGBoost\cite{chen2016xgboost} & 0.802  & 0.814 &0.749 &0.662 &0.782 &0.813 \\
 Adaboost\cite{freund1997decision} & 0.773  & 0.825 &0.763 &0.772 & 0.694 &0.790\\

 \bottomrule
\end{tabular}
  \begin{tablenotes}
        \scriptsize
        \item[a] The Precision and F1-score are the macro averages.  
      \end{tablenotes}
 \end{threeparttable}}
\end{table}

\begin{table}[!t]
\centering

\caption{Performance comparison of different methods on AstraZeneca dataset.}
\label{AZ_set}
\scalebox{1.}{
\begin{threeparttable} 
\begin{tabular}{p{2cm}|p{1.5cm}|p{1.5cm}|p{1.5cm}|p{1.5cm}|p{1.5cm}|p{1.5cm}}
\toprule
\textbf{Metric} & \textit{AU ROC} & \textit{AU PRC} & \textit{ACC} & \textit{BACC} &\textit{Precision}\tnote{a} &\textit{F1-Score} \\
\midrule
 \textbf{Ours}   &\textbf{0.841}  &\textbf{0.887} &0.824 &\textbf{0.858} & \textbf{0.874} & \textbf{0.868} \\
 DeepDDS  & 0.722   & 0.801 &0.654 &0.627 &  0.824 &0.742         \\ 
 DeepSynergy  & 0.681  & 0.726 &0.662 &0.673 & 0.741 &0.735       \\ 
 MR-GNN  & 0.713  & 0.768 &\textbf{0.836} &0.621 & 0.690 &0.704        \\
 MatchMaker  & 0.702 & 0.698 &0.735 &0.728 & 0.790 &0.745        \\
 XGBoost & 0.542  & 0.589 &0.697 &0.596 & 0.623 &0.606        \\
 Adaboost & 0.521  & 0.546 &0.655 &0.620 & 0.633 &0.594\\

 \bottomrule
 \end{tabular}
  \begin{tablenotes}
        \scriptsize
        \item[a] The Precision and F1-score are the macro averages.  
      \end{tablenotes}
 \end{threeparttable}}
 
\end{table}

\begin{table}[t]
\centering
\caption{Performance of different methods on  unseen drug and cell line experiments}
\label{unseen}
\scalebox{1.}{
\begin{threeparttable}

\begin{tabular}{cccc|ccc} 
\toprule
\multirow{2}{*}{Method}     & \multicolumn{3}{c}{Independent Drugs(39)\tnote{a}} & \multicolumn{3}{c}{Independent Cell-Lines(10)\tnote{b}}  \\ 
\cmidrule(lr){2-4}\cmidrule(lr){5-7}
                             & \textit{AU ROC}   & \textit{AU PRC}   & \textit{F1-Score}                & \textit{AU ROC}   & \textit{AU PRC}   & \textit{F1-Score}\tnote{c}                   \\ 

\midrule
\textbf{Ours}   &\textbf{0.834}  &\textbf{0.823} &\textbf{0.854} &\textbf{0.948} & \textbf{0.918} & \textbf{0.963} \\
 DeepDDS  & 0.697   & 0.795 &0.644 &0.889 &  0.854 &0.826       \\ 
 DeepSynergy  & 0.653  & 0.713 &0.676 &0.858 & 0.824 &0.863      \\ 
 MatchMaker  & 0.673  & 0.689 &0.675 &0.863 & 0.894 &0.878        \\
 XGBoost & 0.510 & 0.581 &0.654 &0.794 & 0.754 &0.802       \\
 Adaboost & 0.508 & 0.516 &0.592 &0.746 & 0.703 &0.776\\

\bottomrule
\end{tabular}
\begin{tablenotes}
        \scriptsize
        \item[a] 39 drugs are not included in the training set
        \item[b] 10 cell lines show only a few times (around 10 times) in the training set
        \item[c] The F1-score is the macro average.  
      \end{tablenotes}
\end{threeparttable}
}
\end{table}

\begin{table}[t]
\centering
\caption{Ablation study results}
\label{abalation}

\begin{threeparttable}

\begin{tabular}{cccc|ccc|cccc} 
\toprule
\multirow{2}{*}{Method}     & \multicolumn{3}{c}{DrugComb} & \multicolumn{3}{c}{AstraZeneca} & \multirow{2}{*}{Embedding} & \multicolumn{3}{c}{DrugComb}\\
\cmidrule(lr){2-4}\cmidrule(lr){5-7}\cmidrule(lr){9-11}
                             & \textit{AU ROC}   & \textit{AU PRC}   & \textit{F1-Score}               & \textit{AU ROC}   & \textit{AU PRC}   & \textit{F1-Score}    & &\textit{AU ROC}   & \textit{AU PRC}   & \textit{F1-Score}               \\ 
 
\midrule
\textbf{Ours}   &\textbf{0.961}  &\textbf{0.954} &\textbf{0.972}&\textbf{0.841} &\textbf{0.887} & \textbf{0.868} & -Drug &0.914 &0.920 &0.939\\
 Ours-S\tnote{b}  & 0.946   & 0.925 &0.955 &0.832 &  0.854 &0.887  & -Protein &0.928 &0.915 &0.932\\   
  Ours-P\tnote{c}  & 0.955   & 0.937 &0.943 &0.820 &  0.844 &0.868   & -Disease &0.946 &0.937 &0.952\\  
 Ours-PS\tnote{d}  & 0.943  & 0.936 &0.948 &0.811 & 0.851 &0.837 & -All &0.889 &0.882 &0.913\\    
\bottomrule
\end{tabular}
\begin{tablenotes}
        \scriptsize
        \item[a] The F1-score is the macro average. 
        \item[b] Ours-S means removing self-training strategy from our original model.
        \item[c] Ours-P means removing predictive module (DDI and DTI) from our original model.
        \item[d] Ours-PS means removing both self-training strategy and predictive module from our original model.
      \end{tablenotes}
\end{threeparttable}
\end{table}

\begin{figure}[!t]
\centering
\includegraphics[width=1\textwidth]{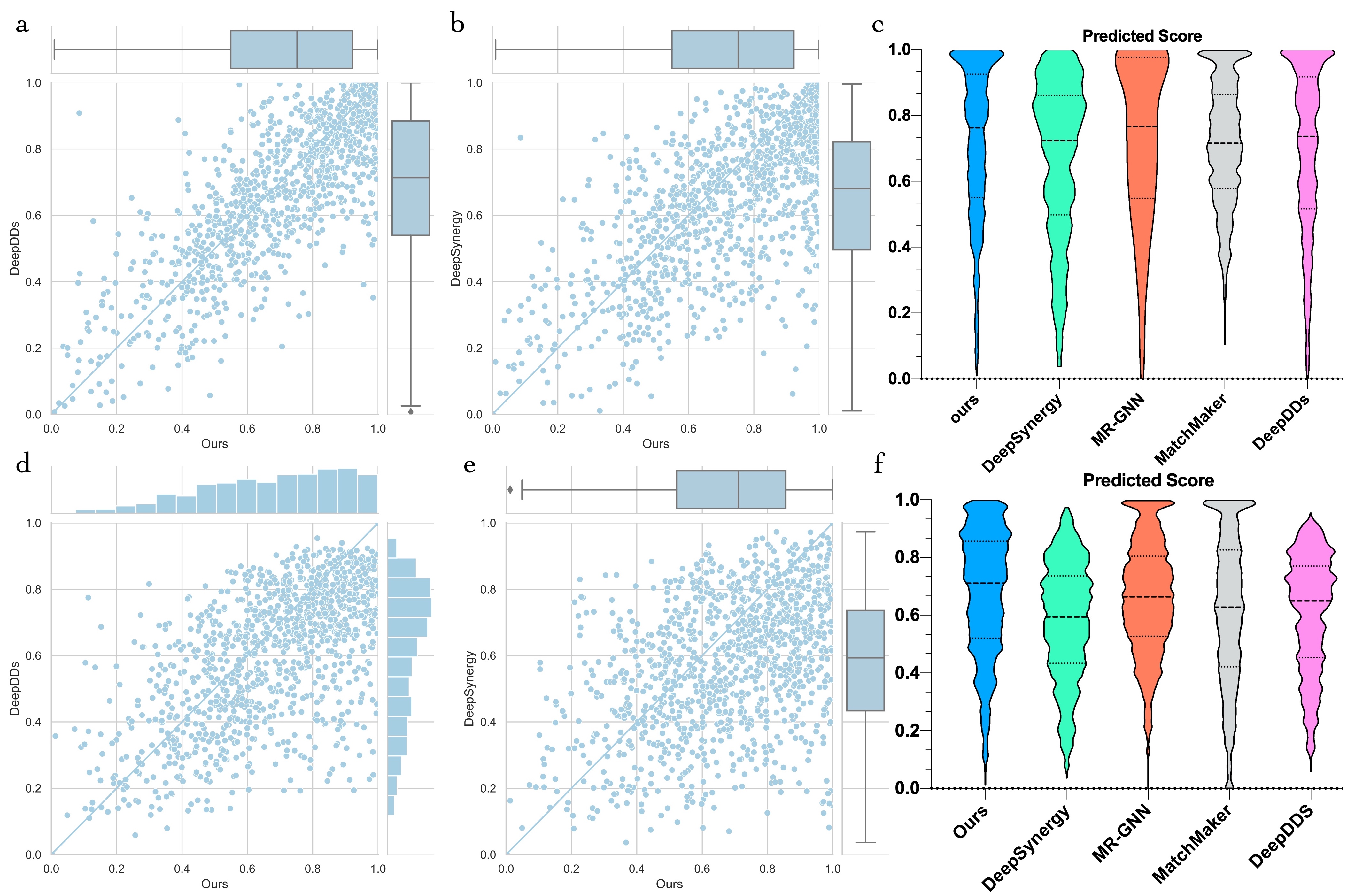} 
\caption{
\textbf{Performance analysis on DrugComb and Astrazeneca.}  \textbf{a.} A scatter plot of predicted scores of Our method and DeepDDS. We measured the confidence scores of our method and DeepDDS for subsampled testing data in DrugComb  (including both synergy and antagonism data). We outperform DeepDDS for more data points are below the diagonal, and our overall predicted score is higher than DeepDDS \textbf{b.} A scatter plot of predicted scores of Our method and DeepSynergy. We measured the confidence scores of our method and DeepSynergy for subsampled testing data in DrugComb. Our model can obviously predict the drug combination effect more accurately than DeepSynergy. \textbf{c.} Violin plots of predicted scores for five models. We compare our method with DeepDDS, DeepSynergy, MR-GNN, and MatchMaker. From the figure, our model's median and quartiles are all higher than the other four methods, indicating that we are among the best.
\textbf{d.} A similar scatter plot of predicted scores of our method and DeepDDS displaying all data points on the independent test set AstraZeneca. We outperform DeepDDS for more data points are below the diagonal, and our overall predicted score is higher than DeepDDS \textbf{e.} A similar scatter plot of predicted scores of Our method and DeepSynergy on AstraZeneca. Our model can obviously predict the drug combination effect more accurately than DeepSynergy. \textbf{f.} Violin plots of our model, DeepDDS, DeepSynergy, MR-GNN, and MatchMaker on AstraZeneca.
}
\label{Fig.DCperformance}
\end{figure}

\begin{figure}[!t]
\centering
\includegraphics[width=1\textwidth]{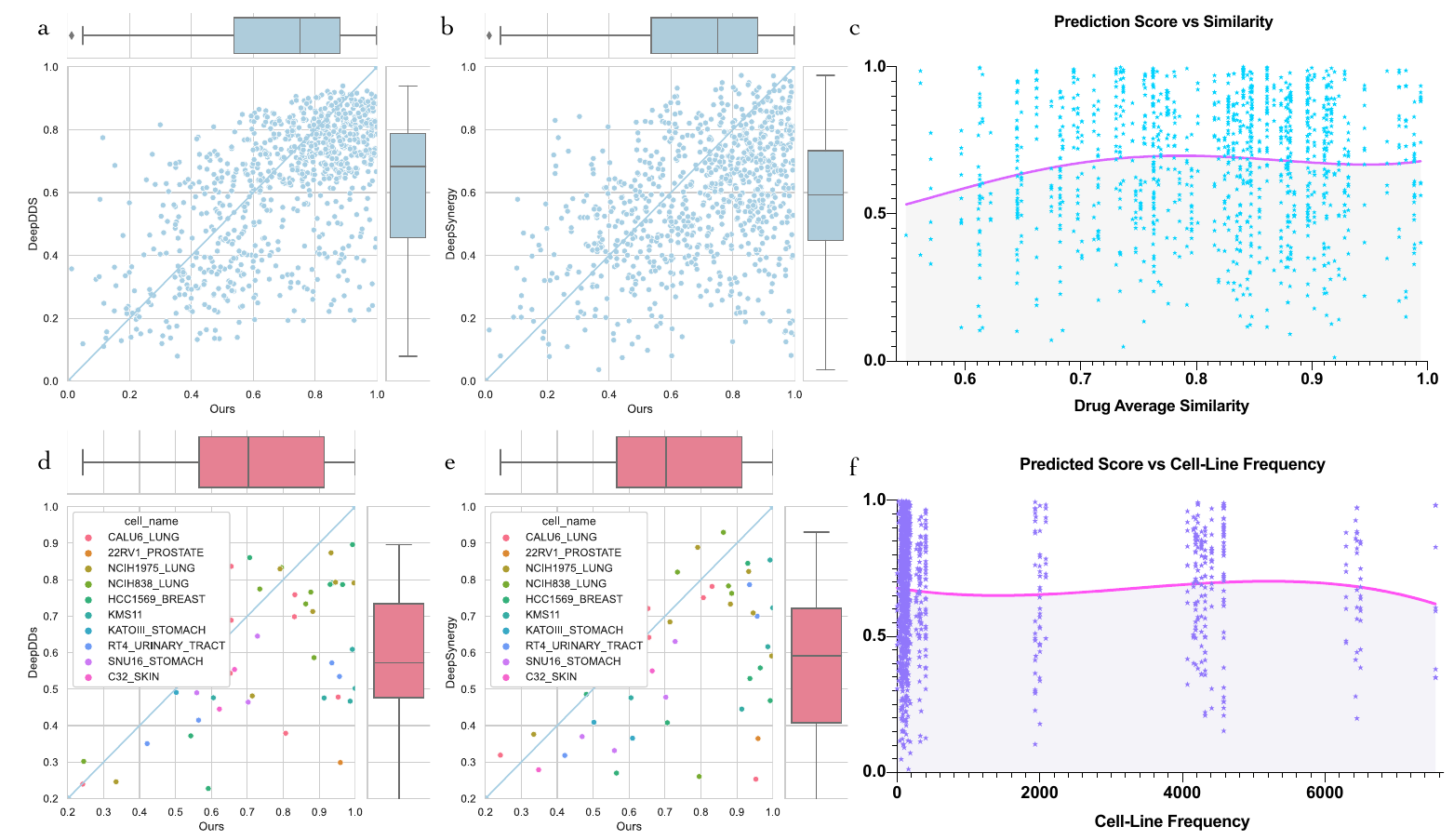} 
\caption{
\textbf{Performance on independent drugs and cell lines.} \textbf{a.} We want to test if our model can accurately predict unseen drugs in the training set. Totally we curated a testing dataset consisting of 39 unseen drugs. Our model surpasses DeepDDS with a higher confidence score. Most data points lie below the diagonal and in the right half of the subgraph. \textbf{b.} A scatter plot of predicted scores of Our method and DeepSynergy. We show more significant results. \textbf{c.} We compare our prediction score against drug similarity. The x-axis expresses the similarity between the drug in test set as well as its most similar drug in the training set, and the y-axis shows the corresponding predicted score of its synergistic effect. We apply Tanimoto Similarity to compute the similarity between the drugs. Although the performance fluctuates slightly, our model behaves robustly. \textbf{d.} 
Here we choose 10 cell lines that hardly ever appear in the training set, and compare the performance of our method with DeepDDS.  \textbf{e.} A scatter plot of predicted scores of our method compared with DeepSynergy. \textbf{f.} We compare our prediction score against cell line frequency. The x-axis refers to the frequency of cell lines in the training set, and the y-axis corresponds to the predicted score for the cell line. Our model shows stable performance across all the appearing frequencies of cell lines}
\label{Fig.celldrugspecific}
\end{figure}

Our model is a fully differentiable end-to-end method to perform Drug synergistic combinations prediction from given drug sequences and cell line representation alone. The overview is illustrated in Figure \ref{Fig.overview}. We performed several techniques such as large-scale pre-training to learn the structural information from a massive amount of unlabeled data, adaptive self-training to enrich our data space, and prediction module to enable inference via pseudo edges on independent nodes. In this section, we present a comprehensive evaluation study and compare our model with other existing methods across multiple metrics: area under the receiver operator characteristics curve \textit{AU ROC}, area under the precision-recall curve \textit{AU PRC}, accuracy \textit{ACC}, balanced accuracy \textit{BACC},  positive predictive value \textit{precision} and the harmonic mean of the precision and recall \textit{F1-Score}.

\paragraph{Cross-validation on DrugComb dataset}

The first step towards analyzing our performance is to compare with state-of-the-art methods on a large benchmark dataset: DrugComb \cite{zagidullin2019drugcomb}. Seven methods are selected including five deep learning methods: DeepDDS, TranSynergy, DeepSynergy, MR-GNN, MatchMaker, and two classical machine learning(ml) methods: XGBoost, and Adaboost.  We utilized ChemicalX \cite{rozemberczki2022chemicalx} reimplementation of these methods except for the official version of DeepDDS and MatchMaker. Their reimplementations are different from the original paper, thus we compared their original and reimplemented versions and reported the best performing results.  While MR-GNN is initially created to predict Drug-Drug Interactions (DDI), we can slightly adapt it to meet our scenarios. For Adaboost or XGBoost, we employed molecular fingerprints as input features of drugs.

The dataset is divided into ten splits of equal size randomly, and we perform cross-validation by iteratively masking out one split for testing and the remaining for training. The value for each metric takes the average numbers across every fold. Detailed results are presented in Table \ref{test_set} and  the top value in each column is highlighted. We achieved the best  result across all metrics except for ACC and precision. Specifically, for the most important measurements AU ROC and F1-Score, we surpass the second-best method DeepDDs by around 2\% and classical ml methods by almost 20\%. Deep learning methods perform better but it is notable that the overall performances for all methods are relatively high on DrugComb, thus the gap between ours and others is not so significant. 

Since this is a binary classification task, we unified the last block of all methods to be a softmax layer. Intuitively, the higher the predicted score in the corresponding correct category, the better the model performs. Thus, we take a step further to examine this prediction score. Figure \ref{Fig.DCperformance} delivers the head-to-head comparison and violin plot of the prediction score.  Figure \ref{Fig.DCperformance}.a and b compare our method with a competitive method: DeepDDS and an Astrazeneca-developed model: DeepSynergy. The x-axis refers to our method while the y-axis for others, each point represents a drug-pair-cell-line data and we subsample them down to 1100 data points. The majority of points are scattered below the diagonal and in the right part of the two subfigures suggesting we gained not only correct labels but also high prediction scores (confidence). Our results are more significant compared to DeepSynergy. Noted that points above the diagonal does not necessarily indicate wrong predictions, while points gathering in the lower left corner indicate hard samples where both methods failed. The boxplot around shows the average prediction scores for the correct category. The violin plot is prediction score distributions for selected four deep learning methods. We achieved the highest prediction score and DeepDDs/MR-GNN achieved similar second-best values. The margin is also not so large on DrugComb dataset.

\paragraph{Accurate predictions on domain-shift dataset}

The sound performance on DrugComb is somehow expected since it contains sufficient training samples and testing is done within the same data domain. Therefore, we tend to evaluate on a domain-shift dataset published by AstraZeneca to examine the performance of cross-domain inference. The data samples within AstraZeneca include plenty of unseen drugs and cell lines thus the drop of performance is no surprise. This time, all methods are then trained on the full DrugComb dataset and conduct inference on AstraZeneca.

Table \ref{AZ_set} summarized our detailed performance on the AstraZeneca dataset, the top result is highlighted in each column, and overall degraded performance is observed. Particularly, our \textit{AU ROC} fall from 0.961 to 0.852 and F1-Score from 0.972 to 0.863 compared with DrugComb test. This may be due to the data domain-shift but noticed that the second best DeepDDS or MR-GNN drop remarkably from 0.94 to 0.72 regarding their \textit{AU ROC}. The margin of our model with DeepDDs enlarges from 0.02 to 0.14, this is a substantial improvement since real application scenarios always exist data domain-shifts and new drugs. Thus, from the result, we can reduce more than 14 wrong cases out of 100 testing samples than DeepDDS, which can save great human effort and time costs in drug discovery. The results in terms of F1-Score behave similarly as our model performs better on imbalanced data.
We conduct a similar analysis of the prediction score as it is in cross-validation, Figure \ref{Fig.DCperformance} again delivers the head-to-head comparison and violin plot of prediction scores. Figure \ref{Fig.DCperformance}.d and e compares our method with two typical models in drug combination effect prediction methods: DeepDDs and DeepSynergy. The x-axis refers to our method while the y-axis for others. There are around 1200 datapoints in AstraZeneca, Majority of them are more scattered below the diagonal and in the right part of the two subfigures compared to DrugComb, suggesting that we're  achieving more significant results on AstraZeneca dataset, which are not only correct labels but also high prediction scores (confidence). The box-plot around shows the average prediction score for the correct category. The violin plot in Figure \ref{Fig.DCperformance}.f shows prediction score distributions for four deep learning methods. We achieved the highest prediction score and DeepDDs/MR-GNN again achieved a similar second-best value.

\paragraph{Inferencing on independent drugs and cell lines}

The results from the AstraZeneca dataset indicate that domain-shift data is hard to infer. In fact, the AstraZeneca dataset still contains overlapping drugs or cell lines with DrugComb. To further test the generalization ability of our model, we tend to study our performance on non-overlapping drugs and cell lines. First, we create two datasets consisting of 39 independent drugs and 10 independent cell lines with 946 and 59 entries, respectively. Noted that these independent data are appointed from Astrazeneca on top of a cross-domain setting, and we delete the selected entries from our training set and the remaining data are used for training purposes. The independent drugs and cell lines are hence unrecognized to our model. Though these tasks are tough for the prior methods, according to our inference step, these unseen drugs can be linked in the graph via generated drug similarity edges or pseudo DTI or DDI edges. In this way, we enable information propagation even on unseen drugs and could provide richer representations than the initial embeddings. Notice that when a drug is dissimilar to all other drugs in datasets and has no predicted DTI/DDI interactions, our model would somehow degrade to an MLP-like classification pipeline, which we will investigate further.

Table \ref{unseen} summarized our performance in these two independent scenes. In the unseen drug study, our method still maintains a relatively high \textit{AU ROC} and \textit{AU PRC} over 80\% while the performance of other methods like DeepDDs and DeepSynergy drop under 70\% in some criteria. We achieved over 85\% regarding \textit{F1-Score} which is 20\% better than DeepDDS. From Figure \ref{Fig.celldrugspecific}.a and b, it can be clearly observed that most data points are below the diagonal and at the right side, such result indicates that our model is more robust against predicting unseen drugs than DeepDDS and DeepSynergy. Figure \ref{Fig.celldrugspecific}.c shows the trend between predicted score and drug similarity. Here the similarity implies the Tanimoto Similarity between  unseen drugs and their most similar drugs in training set. Although the performance drops at low similarity region, our model is able to precisely predict the data above $0.5$, which in the sense conducting correct classification. Also, the regression trend shows our robust performance on the whole.

In the unseen cell line experiment, all the methods obtained a relatively high performance, as shown in Table \ref{unseen} since they all applied CCLE cell line expression as embedding vectors. However, we're different in the sense that we express the cell line as a weighted sum of rich protein embeddings while they treat protein as a one-hot vector. Thus, our model is superior to others for at least 5\% across all criteria. Figure \ref{Fig.celldrugspecific}.d and e present an intuitive view of unseen cell line study results of our model, DeepDDS, and DeepSynergy. Take HCC1569\_BREAST as an example, for most data points of this cell line, our model predicts them with very high confidence, on the contrary, DeepDDS and DeepSynergy can hardly make the right decision. The box plots around also show that our model has a significant lead on overall predicted scores. Besides, we calculated the trend between the predicted score and the occurrence frequency of the cell line in the training set, which is demonstrated in Figure \ref{Fig.celldrugspecific}.f. Our model behaves stably against the changes in cell line frequency. To some extent, the two data perturbation experiments verify that our model maintains high capability on challenging tasks.

\paragraph{Ablation study}
With the aforementioned experiments validating the strengths of our model in predicting drug combination effect, we first conduct an ablation study to evaluate the effectiveness of our submodules: self-training strategy and pre-trained DDI and DTI respectively. To investigate the self-training strategy, we remove it from our model and then regard this pipeline as Ours-S. Furthermore, to analyze the two predictive modules, we banned them from our framework and similarly named this pipeline as Ours-P. At last, we skip both self-training strategy and predictive modules to obtain the ‘Ours-PS’ pipeline. We conduct ablation studies on both DrugComb and AstraZeneca.  Table \ref{abalation} shows all four models delivered high \textit{AU ROC}, \textit{AU PRC}, and \textit{F1-Score}, our original model has been consistently ranked as the best performance across all the measures. Results indicate that our self-training strategy and predictive modules nicely mine more useful information about drugs and proteins, and it helps our model to behave better in drug combination effect prediction.

Meanwhile, we found that even with our basic model ‘Ours-PS,’ we achieved noteworthy results on DrugComb, which could be attributed to our innovative use of rich embeddings over all instances. Therefore, we investigate how performance varies when moving to simpler representations. We used molecular fingerprints as an alternative for drug representations (‘-Drug’), one-hot encoding for either protein (‘-Protein’) or Disease alternative representations (‘-Disease’). ‘-All’ denotes the model using all the above mentioned simple alternative representations. The right part of Table \ref{abalation} concludes that instance embeddings have a huge and explicit impact on our performance. While drug embeddings are closely related to synergistic combinations, the term ‘-Drug’ yields the poorest results. One-hot encoding of Protein and Disease would introduce more noise and less information into our pipeline and thus lower our performance. When all rich representations are deleted, ‘-All’ definitely yields the worst results.

\paragraph{Hyperparameters setting} We listed the model parameters we used in our pipeline here to offer a better understanding. Every layer of our graph neural network is based on graph attention nets with input and output of both 512 dimensions and increased heads 4,8,12. The MLP prediction for synergistic classification has hidden layers of 3072, 768, and 128 dimensions. For DTI and DDI prediction modules, we use 1 attention block with 8 heads to encode either protein or drug representations and 2 attention blocks with 12 heads to  process their concatenated embeddings. An MLP with 2048 and 256 dimensions of hidden layers is followed to predict the outcomes. We trained with a learning rate of $10^{-4}$ and a dropout rate of 0.2.

\section{Conclusion}
We develop an end-to-end model to facilitate the detection of drug combinations, aggregating various types of drug-related information. Comprehensive experiments including cross-fold validation, independent test, ablation study, and unseen experiment, demonstrate the effectiveness and robustness of our proposed method, where our model consistently and significantly outperforms all counterparts. Most known drug combination prediction methods utilize one or two kinds of features and their discovery ability are limited to only a few cell lines or tissues, they are not able to handle novel drugs, leading to poor performance in our unseen experiment. 

Moreover, since large-scale pre-trained models show brilliant results in many fields, extending our model to do multi-tasks is a promising research direction. In the future, we will try to develop a method that is able to conduct multiple drugs, cell lines, and disease-related tasks, not limited to drug combination effect prediction.

We believe that our model can serve as a powerful tool to facilitate precise combination medicine and novel combination discovery. We will try to incorporate other kinds of information, such as 3D molecular structure into our framework to further improve our method’s performance.

\section{Acknowledgement}
The research reported in this publication was partially supported by Research Grants Council of the Hong Kong Special Administrative Region, China (Research Impact Fund (RIF), R5034-18, CUHK 2410021).

\bibliographystyle{unsrt}
\bibliography{my}

\end{document}